\pdfoutput=1

\documentclass[11pt]{article}

\usepackage[]{acl}

\usepackage{times}
\usepackage{latexsym}

\usepackage[T1]{fontenc}

\usepackage[utf8]{inputenc}

\usepackage{microtype}
\usepackage{multirow}
\usepackage{amsmath}
\usepackage{graphicx}
\usepackage{xr}
\usepackage{xspace}
\usepackage{subfigure}
\usepackage{makecell}
\usepackage{threeparttable}
\usepackage{tablefootnote}

\newcommand{\method}{Lasformer\xspace}

\title{Only 5\% Attention Is All You Need: Efficient Long-range Document-level Neural Machine Translation}

\author{
    Zihan Liu\textsuperscript{\rm 1,2\thanks{* Work was done while Z. Liu was an intern at ByteDance.}},\quad
    Zewei Sun\textsuperscript{\rm 2},\quad
    Shanbo Cheng\textsuperscript{\rm 2},\quad
    Shujian Huang\textsuperscript{\rm 1},\quad
    Mingxuan Wang\textsuperscript{\rm 2} \\
    \textsuperscript{\rm 1} National Key Laboratory for Novel Software Technology, Nanjing University \\
    \textsuperscript{\rm 2} ByteDance \\
    \texttt{liuzh@smail.nju.edu.cn}, ~\texttt{huangsj@nju.edu.cn} \\
    \texttt{\{sunzewei.v,chengshanbo,wangmingxuan.89\}@bytedance.com}
}

\begin{document}
\maketitle
\begin{abstract}
Document-level Neural Machine Translation (DocNMT) has been proven crucial for handling discourse phenomena by introducing document-level context information. One of the most important directions is to input the whole document directly to the standard Transformer model. In this case, efficiency becomes a critical concern due to the quadratic complexity of the attention module. Existing studies either focus on the encoder part, which cannot be deployed on sequence-to-sequence generation tasks, e.g., Machine Translation (MT), or suffer from a significant performance drop. In this work, we keep the translation performance while gaining 20\% speed up by introducing extra selection layer based on lightweight attention that selects a small portion of tokens to be attended. It takes advantage of the original attention to ensure performance and dimension reduction to accelerate inference. Experimental results show that our method could achieve up to 95\% sparsity (only 5\% tokens attended) approximately, and save 93\% computation cost on the attention module compared with the original Transformer, while maintaining the performance.
\end{abstract}

\section{Introduction}
Recent developments in neural machine translation have focused on the translation of individual sentences, but research has shown that document-level information is crucial for handling discourse phenomena such as lexical consistency and pronominal anaphora, which rely on long-range context. As a result, various attention mechanisms~\cite{DBLP:conf/emnlp/ZhangLSZXZL18,DBLP:conf/naacl/MarufMH19,DBLP:conf/ijcai/ZhengYHCB20,DBLP:conf/acl/Bao0TCL20} that encode document-level context information have been proposed. 

However, the computation cost of these attention mechanisms increases quadratically with the length of the input sequence. To address this issue, researchers have proposed efficient transformer models~\cite{DBLP:journals/corr/abs-2009-06732/survey} that aim to reduce the computation cost of attention through techniques such as sparsity patterns~\cite{DBLP:conf/icml/TayBYMJ20,DBLP:journals/corr/abs-1904-10509/Sparse,DBLP:conf/nips/ZaheerGDAAOPRWY20,DBLP:journals/corr/abs-2004-05150/Longformer} that limit the number of tokens to attend to, memory or global tokens that compress contextual tokens into a single representation~\cite{DBLP:conf/icml/LeeLKKCT19,DBLP:conf/nips/MaKWZMMZ21}, approximation to softmax with kernel methods~\cite{DBLP:journals/corr/abs-2009-14794/Performers,DBLP:conf/iclr/QinSDLWLYKZ22,DBLP:conf/iclr/Peng0Y0SK21}, or a combination of above~\cite{DBLP:conf/icml/TayBMJZZ21,DBLP:conf/nips/ZhuPXSGAC21}.

Despite the emergence of various efficient transformer models, long-range sequence-to-sequence tasks such as document-level machine translation still need more exploration.

On the one hand, some of the existing efficient models~\cite{DBLP:journals/corr/abs-2006-04768/Linformer,DBLP:conf/nips/ZaheerGDAAOPRWY20,DBLP:conf/naacl/Lee-ThorpAEO22} focus on the encoder part and can not be used for generation because of the auto-regressive property. Some~\cite{DBLP:conf/iclr/Tay0ASBPRYRM21,DBLP:journals/corr/abs-1904-10509/Sparse,DBLP:journals/corr/abs-2004-05150/Longformer} have a strong relationship to the position of tokens thus can not be applied to cross attention where no alignment is obvious between query and key. 

On the other hand, the studies that target on efficient sequence-to-sequence generation only verify their methods on normal sentence-level translation benchmarks like WMT EN-DE test sets~\cite{DBLP:conf/iclr/Peng0Y0SK21,DBLP:conf/iwslt/PetrickRHN22,DBLP:conf/nips/MaKWZMMZ21}. In our preliminary experiments, we find that almost all the work severely drops in BLEU when dealing with real document translation tasks. 



To address this issue, we try to reduce the computation cost while ensuring the translation performance.
In this paper, we mainly focus on the attention mechanism following other efficient transformer models.

Specifically, we want to select important tokens~\cite{DBLP:conf/aaai/SunHWDC20,DBLP:conf/coling/SunHDC22} and only conduct attention to them.
Previous studies sharing a similar motivation either design sparsity patterns with human prior like a fixed sliding window\cite{DBLP:journals/corr/abs-2004-05150/Longformer,DBLP:conf/icml/TayBYMJ20,DBLP:conf/nips/ZaheerGDAAOPRWY20} which lack flexibility, or try to learn the sparsity pattern by clustering methods. However, the poor performance of learnable pattern methods on DocNMT reflects that the query does not attend to the keys expected in original attention. 

In order to ensure the performance, we take advantage of the original attention and propose Lightweight Attention Selection Transformer (\method ). \method incorporates selection layers that utilize lightweight attention, whose distribution is guided by supervision from the original attention. The achievement of lightweight processing is attained by reducing the hidden dimension, while the selection process involves retaining tokens with the highest attention scores, a strategy validated for its efficacy by \cite{zhao2019explicit}. By employing these mechanisms, we are able to efficiently filter out insignificant tokens at a comparatively low expense, resulting in a reduction of the overall computational burden, particularly when a significant proportion of tokens can be filtered out.


Determining the appropriate number of tokens to retain is of utmost importance, as they must contribute sufficient information to ensure optimal performance, while also minimizing their quantity to enhance efficiency. In our approach, the sparsity is learned adaptively, which gradually increases during the training process until it reaches an optimal level that strikes a balance between performance and efficiency for each selection layer.



Experiments show that \method can effectively reduce the computation of attention. Only 5\% of tokens are used in attention and translation performance remains almost unchanged. For the long sequence of thousands of words, our method can lower the attention cost to 7\%. And end-to-end inference speed can be enhanced to 1.2x.

\section{Related Work}
\begin{figure*}[htbp]
    \centering
    \begin{minipage}[t]{0.4\textwidth}
        \centering
        \includegraphics[width=1.0\textwidth]{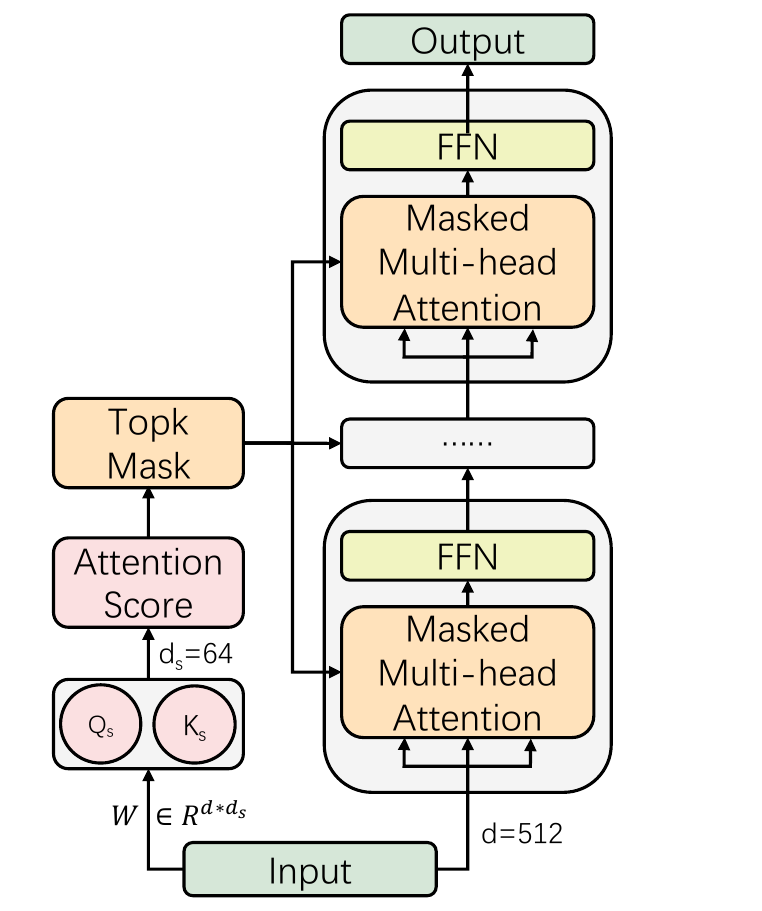}
    \end{minipage}
    \begin{minipage}[t]{0.57\textwidth}
        \centering
        \includegraphics[width=1.0\textwidth]{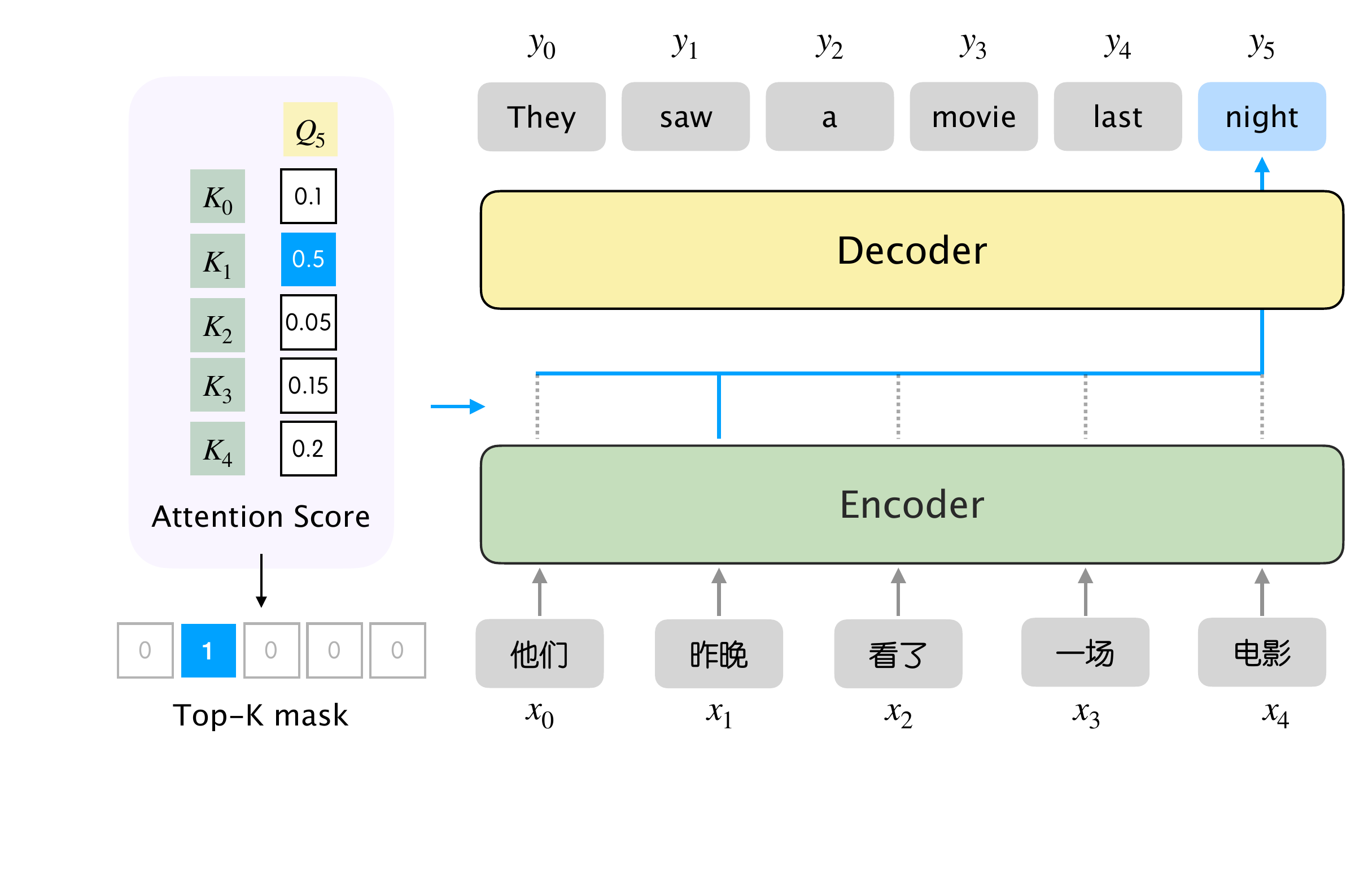}
    \end{minipage}
    \caption{The left figure is the whole architecture of \method.
    Its left part represents the selection module. It accepts a low-dimensional input to calculate lightweight attention. With the rough attention, we mask the unimportant tokens in the main module (right part of the left figure). In addition, the selection mask is shared across some layers. The right figure illustrates the masking procedure. Taking cross-attention as an example, the current query only attends to those tokens with high rough attention.}
    \label{fig:architecture}
\end{figure*}

\subsection{Document-level Machine Translation}
Document-level machine translation involves an additional source and target context to improve the translation in terms of coherence and consistency~\cite{voita-etal-2019-good,muller-etal-2018-large,lopes-etal-2020-document,bawden-etal-2018-evaluating}. There exist two lines of methods to use context. One introduces an extra encoder to encode context and integrate it into the current sentence~\cite{DBLP:conf/emnlp/ZhangLSZXZL18,DBLP:conf/naacl/MarufMH19}. The limitation is that the same sentence might be encoded multiple times thus increasing the complexity. It is solved by recent works by sharing the parameters of context encoder and current sentence encoder~\cite{DBLP:conf/ijcai/ZhengYHCB20,ma-etal-2020-simple}.

Another line of work concatenates the context and the current sentence and translate it as if it is a single sentencec~\cite{tiedemann-scherrer-2017-neural,DBLP:conf/acl/SunWZZHCL22}. However, the concatenation results in a long input sequence and makes it difficult to train the model, because of the high entropy of attention distribution. To alleviate the problem, locality bias is introduced, where sentence-level information is augmented~\cite{DBLP:conf/acl/Bao0TCL20}. 

In short, the former is based on sentence translation while integrating the context. The latter tries to translate the whole document while introducing sentence-level locality. And they seem to reach the same scheme that uses both local attention and global attention.

The local attention implies human-designed sparsity pattern and it is natural to introduce learnable sparsity pattern to global attention in document-level machine translation.


\subsection{Efficient Transformer}
There have been several previous methods for efficient Transformers that have focused on the properties of attention, specifically sparsity and low rank to reduce the computation cost. 

Sparsity refers to the idea that only a few tokens receive a significant amount of attention, while the rest contribute little to the output. Some methods~\cite{DBLP:conf/icml/TayBMJZZ21,DBLP:journals/corr/abs-1904-10509/Sparse,DBLP:journals/corr/abs-2004-05150/Longformer,DBLP:conf/nips/ZaheerGDAAOPRWY20} have proposed handcrafted patterns such as the sliding window or dilated window, which is inspired by human prior knowledge that close tokens contribute the most attention. Other methods~\cite{DBLP:conf/iclr/KitaevKL20,wang-etal-2022-clusterformer,DBLP:conf/icml/TayBYMJ20,roy-etal-2021-efficient} have attempted to make the sparsity pattern learnable with a lower cost by using techniques like clustering, based on the idea that similar tokens are expected to attend to each other and belong to the same cluster. These clustering methods can include techniques like locality sensitive hashing~\cite{DBLP:conf/iclr/KitaevKL20}, K-means~\cite{roy-etal-2021-efficient}, or learnable sorting networks~\cite{DBLP:conf/icml/TayBYMJ20}. 

On the other hand, low-rank methods are based on the idea that $N$ dimensional features can be compressed into fewer dimensions.
Some work~\cite{DBLP:conf/icml/LeeLKKCT19,DBLP:conf/nips/MaKWZMMZ21,DBLP:conf/icml/JaegleGBVZC21} has used global tokens or memory to compress long-range information into a limited number of embeddings or has used kernel methods~\cite{DBLP:conf/iclr/Peng0Y0SK21,DBLP:conf/iclr/QinSDLWLYKZ22} to approximate softmax scores, allowing the computation of keys and values first and reducing the complexity from $O(N^2d)$ to $O(Nd^2)$ (where $d$ is the dimension of self-attention).

While sparsity methods maintain a token-to-token attention structure, low-rank methods use a compressed embedding for attention. The token-to-token approach is more interpretable but may lose some information, while the other may contain more information but may also be noisier. Since the information in DocNMT is sparse\cite{lupo2022divide}, the noise of low-rank methods might be much more severe and thus we exploit the sparsity methods. 


\section{Method}

Sequence-to-sequence document-level translation aims at fully capturing the distant context. It is achieved by attention mechanism, which allows each query attending to all of the keys, resulting in quadratically growing computation cost with the sequence length. However, only a quite small part of tokens is truly relevant. Therefore, it is important to select those important ones and filter the others to reduce the number of aligned objects. 


Specifically, we take advantage of the origin attention mechanism and distill it into a lightweight attention with lower hidden dimension to select important tokens, as shown in Figure~\ref{fig:architecture}. It still takes O$(N^2)$ calculation but has much less computation. After filtration, only remaining tokens will be attended. Although the selection introduces extra cost, the total efficiency can be improved as far as the aligned range is limited enough.

Basically, we divide our methods into four parts: lightweight attention, attention supervision, adaptive sparsity, and layer sharing, which will be introduced in the following sections.

\subsection{Lightweight Attention}
Suppose the sequence has $N$ tokens in total and we need to select $kN$ tokens that are important for the current token. $k$ is the selection ratio. Since the selection is only a preliminary process and should only take very little calculation, we project the hidden state of all the tokens from $d$ to $d_s$ (e.g. from 512 to 64). Then a lightweight version of attention is conducted with those low-dimensional hidden states:

\begin{equation}
    A_s = \text{softmax}(Q_sK_s^T/\sqrt{d_s}) \\ 
\end{equation}

where $Q_s$, $K_s$, and $A_s$ represent the projected query, key, and attention.
$Q_s = XW_Q$, $K_s = XW_K$, and $W_Q \in \mathbf{R}^{d \times d_s}$, $W_K \in \mathbf{R}^{d \times d_s}$.

After sorting all the logits, we only preserve the top $k$ keys for each query token and mask the others:

\begin{equation}
    mask = \text{top-k}(A_s) \\
\end{equation}

Obviously, the top-k function is not differentiable. To train the selection network, we use the re-parameter trick from Gumbel Softmax\cite{DBLP:conf/iclr/JangGP17} to make the parameters learnable:

\begin{equation}
    mask = mask + A_s - SG(A_s) \\
    \label{trick}
\end{equation}

where SG refers to stop-gradient. Then the gradient can be passed to $A_s$ while remaining the value of the mask.

\subsection{Attention Supervision}

Intuitively, the distribution of lightweight attention should be consistent with the original attention layer to ensuring performance. Therefore, we pulls the former to the latter during the training by an addition KL loss. Such distilling process requires no pretrained Transformer model, but the low and high dimension layers are trained with consistency constraint. However, the utilization of original attention prevent speeding up at training time, so we only focus the inference efficiency.

\begin{equation}
    A = \text{softmax}(QK^T/\sqrt{d}) \\
\end{equation}
\begin{equation}
    L_s = \text{kl\_div}(A_s, A) \\ 
\end{equation}

where $Q$ and $K$ are high-dimensional projected hidden states from the original attention layer. kl\_div is the Kullback-Leibler Divergence. The loss is added to the NMT loss with a hyper-parameter $\alpha$:
\begin{equation}
    Loss = L_{nmt} + \alpha * L_s \\
    \label{loss}
\end{equation}

\subsection{Adaptive Sparsity}

$k$ represents the level of sparsity and is important in the whole selection procedure. However, the optimal choice of $k$ is not apparent. We propose an adaptive algorithm to search for it. 

Specifically, we set a threshold $t$, for the sum of attention. The intuition is that, since a small amount of tokens contribute to most of the attention weights, ``the most of weights'' can be quantified as threshold $t$. If the current sum of attention is below $t$, some important tokens might be filtered, so we slightly increase $k$ for a small step, and vice versa:


\begin{equation}
    k = \begin{cases}
            k - step &\text{if sum(topk) > t}\\
            k + step &\text{else}
        \end{cases}
\end{equation}

We regard $k$ as a percentage, so $k$ is in the range $[0,1]$, and the step is a small constant such as $0.001$. We initialize $k$ as 1 and limit $k$ great than or equal to 1\%. For documents with few sentences, at least 10 tokens are attended to avoid poor performance. While $k$ gradually decreases and converges in the training process, the model is encouraged to learn a concentrated attention distribution and get rid of unrelated information. In some layers, especially encoder layer, $k$ might stuck at some point and sometimes never decrease, so we manually disable $k + step$ when $k$ is large.





\subsection{Layer Sharing}


Furthermore, we share the learned sparsity patterns across layers as \cite{xiao2019sharing} has proved that attention weights can be directly reused because intuitively, each query in different layers often attends to the same keys. So the extra selection cost can be further reduced while keeping the translation performance. 

Basically, we divide all the selection layers into $m$ groups and each group has $r=n/m$ layers, where $n$ is the original layer number.
We only calculate the attention of the lowest selection layer in each group. Then the other selection layers share the same attention as the lowest one:

\begin{equation}
    A_{s_i} = A_{s_{\lfloor i/r \rfloor *r}}
\end{equation}

In this way, we can save $m*(r-1)$ calculation of attention selection.

\subsection{Cost Saving}

In the end, we try to formalize the attention cost with these algorithms and parameters. The attention cost of the original Transformer attention~\cite{DBLP:journals/corr/abs-2009-06732/survey}:


\begin{equation}
    \begin{aligned}
        C_{Transformer} = 2nN^2d
    \end{aligned}
\end{equation}

where $n$ is the layer number, $N$ is the sequence length, and $d$ is the dimension of the hidden states. 
``2'' means dot product ($A=QK$) and weighted sum ($AV$).
And \method can achieve a complexity as:


\begin{equation}
    \begin{aligned}
        C_{\method} &= \frac{1}{r} \cdot nN^2 d_s + 2knN^2d
    \end{aligned}
    \label{eq:comlexity}
\end{equation}

where $r$ is the layer number in each group, $d_s$ is the dimension of the selection layer, and $k$ is the selection ratio. The first item means the $d_s$-dimension rough selection. The second item means masked attention.



With a small dimension for selection ($d_s$), a high sparsity for attention ($k$), and a large layer group size ($r$), we can greatly reduce the total computation cost.
If we set $n=6$, $d=512$ as Transformer base, and $d_s=64$, $t=0.95$ ($k=0.05$), $r=3$, the attention cost can be only \underline{\textbf{7\%}} compared with original Transformer. The detailed results are listed in the following sections.


\begin{table*}[htbp]
    \centering
    \begin{threeparttable}[b]
        \begin{tabular}{l|cc|cccc}
            \Xhline{2\arrayrulewidth}
             & \multicolumn{2}{c|}{Efficiency} & \multicolumn{4}{c}{Quality (BLEU)} \\
            \multicolumn{1}{c|}{Models} & Attn Cost & Infer Speed & TED                   & News                  & Europarl              & PDC   \\ \hline
            \tnote{$\heartsuit$} ~~Transformer~\cite{DBLP:conf/acl/SunWZZHCL22}  &  100\% & 1.0x & 27.96 & 25.05 & 34.48 & 27.80 \\ 
            \tnote{$\spadesuit$} ~~LSH-trans~\cite{DBLP:conf/iwslt/PetrickRHN22}        &  2\% & 0.8x & 9.80 & 10.04 & 18.44 & 17.82      \\
            \tnote{$\diamondsuit$} ~~Luna~\cite{DBLP:conf/nips/MaKWZMMZ21}              & 12\% & 1.5x & 10.15 & 9.02 & 20.32 & 19.44       \\ 
            \tnote{$\clubsuit$} ~~RFA-trans\cite{DBLP:journals/corr/abs-2210-08431/rfa-translation} & 10\% & 1.8x & 16.93 & 16.92 & 26.91 & 23.48 \\
            \method        & 7\%  & 1.2x  &27.24                 & 25.95                 & 34.62                 & 28.04       \\ 
            \Xhline{2\arrayrulewidth}
        \end{tabular}
        \begin{tablenotes}
            \scriptsize
            \item [$\heartsuit$] It adopts the original Transformer and we use MR Doc2Doc setting.
            \item [$\spadesuit$] Its complexity is $nCd(N/C)^2+ndNlogN$, where C is hashing chunk size, and we set C=N/32 as decribed in the paper. The hashing (independent of attention) and sorting take a very long time, yielding a overall low speed.
            \item [$\diamondsuit$] Its complexity is $2nNdm$, where m is the number of compressed tokens. We set m=64.
            \item [$\clubsuit$] Its complexity is $2n(Nd'^2+Ndd')$, where $d'$ is the projection dim set to 128, n is the number of layers and N is the sequence length. We set n=6 and N = 1000 for the above settings.
        \end{tablenotes}
    \end{threeparttable}
    \caption{The results of document-level translation. Except for baseline, we also list the attention cost and inference speed of three typical studies on efficient seq2seq generation. They achieve better efficiency, in terms of cost or speed. However, they face severe drop when dealing with the real document-level translation. Overall, \method achieves the best results of long-range document-level translation.}
    \label{tab:main_table}
\end{table*}

\section{Experiments}
\subsection{Datasets}
We conduct experiments on three English-German datasets and one Chinese-English datasets. The English-German datasets include TED, News, and Europarl, following ~\citet{DBLP:conf/naacl/MarufMH19}. The TED corpus is from the IWSLT 2017, and we use tst2016-2017 as test est and the rest are used for development. News are aligned document-delimited News Commentary-v11 corpus, and WMT’16 newstest2015 and news-test2016 are used for development and testing, respectively. Europarl is extracted as proposed in ~\citet{DBLP:conf/naacl/MarufMH19}. For Chinese-English datasets, we follow ~\citet{DBLP:conf/acl/SunWZZHCL22}, using PDC  which is crawled bilingual news corpus with diverse domains.

The above training data are organized into a mix of sentence-level data and document-level data as used in ~\citet{DBLP:conf/acl/SunWZZHCL22}. All of the data are cut into sub-words using BPE with 32k merge operations.


\subsection{Model settings}
We build our translation model based on Transformer base~\cite{DBLP:conf/nips/VaswaniSPUJGKP17} using fairseq\cite{ott2019fairseq}, including 6 layers, 512 dimensions, 8 heads, 2048 feed-forward hidden size, for both encoders and decoders. We use a small dropout of 0.1, as well as word dropout, on large datasets like Europarl and PDC, and a large dropout of 0.3 on small datasets like TED and News.

As for our proposed selection layer, we use $d_s=64$ dimensions, $m=2$ groups, and $r=3$ layers.
The coefficient $\alpha$ is set to 0.01 and the threshold $t$ for dynamic top-k is set to 0.95.

We adopt case-insensitive sacreBLEU~\cite{post-2018-call/sacreBleu} on the whole documents, following all the previous document-level NMT studies.

\subsection{Comparison Work}

We compare the results with three typical efficient Transformers from different classes of methods and directly use their open-source code to conduct experiments on the datasets:

\begin{itemize}
    \item \textbf{LSH-trans}~\cite{DBLP:conf/iwslt/PetrickRHN22} \footnote{\url{https://github.com/rwth-i6/returnn-experiments/tree/master/2022-lsh-attention}} is based on Reformer and uses locality sensitive hashing to obtain a cluster of tokens to be attended to each other within it. 
    \item \textbf{Luna}~\cite{DBLP:conf/nips/MaKWZMMZ21}\footnote{\url{https://github.com/XuezheMax/fairseq-apollo}} is a low-rank-based model that compresses the long sequence into a fixed number of global tokens using the attention mechanism.
    \item \textbf{RFA-trans}~\cite{DBLP:journals/corr/abs-2210-08431/rfa-translation} extends the RFA~\cite{DBLP:conf/iclr/Peng0Y0SK21} with sentence level gating mechanism to enhance the locality\footnote{\url{https://github.com/ZhaofengWu/rfa-doc-mt}}.
\end{itemize}

There are many other efficient Transformer studies\cite{DBLP:journals/corr/abs-2004-05150/Longformer,DBLP:conf/nips/ZaheerGDAAOPRWY20,DBLP:conf/icml/TayBYMJ20,DBLP:conf/icml/TayBMJZZ21}. However, since they bypass sequence-to-sequence generation tasks and only focus on the encoder-only or decoder-only task, we do not involve them here.

\subsection{Results}

Table~\ref{tab:main_table} shows the translation results compared to previous document-level translation models. As for efficiency, all related studies achieve cost saving to various extents. 
They yield better results in terms of cost or speed.
However, they face a serious quality drop when dealing with real long-range documents. We find that although they report a comparable result on WMT or IWSLT (with very limited context, around 30 tokens per sentence), there is a large performance decrease on long documents like TED, Europarl, and PDC. These results are obtained by their open-source codes. We suggest that all efficient-related studies should be verified on real long-range sequences. Otherwise, some potential risks may be ignored. 

Overall, \method achieves the best results, not only reducing the attention calculation and boosting end-to-end inference speed effectively but also maintaining the translation quality. Notably, we cut down the attention cost to 7\%, which is important for the quadratic growth with the sequence length.

\begin{table}[h]
\renewcommand\arraystretch{1.3}
    \centering
    \begin{tabular}{lcccc}
        \Xhline{2\arrayrulewidth}
          Model    & TC    &  CP   & PT & TCP         \\ \hline
          Transformer    &   56.3    &   38.1     & 40.2 &  44.1       \\
          \method    &   54.4    &   37.4     & 41.9  & 44.0           \\
          \Xhline{2\arrayrulewidth}
    \end{tabular}
    \caption{Results on TCP.}
    \label{tab:das_tcp}
    
\end{table}

Meanwhile, except for BLEU, we conduct experiments on document-level test set to evaluate the capability of utilizing document context. We do not use contrastive test sets\cite{voita-etal-2019-good,bawden-etal-2018-evaluating} because their instance only contains at most 5 sentences. Instead, we test our model on PDC\cite{DBLP:conf/acl/SunWZZHCL22}, including Tense Consistency(\textbf{TC}), Conjunction Presence(\textbf{CP}), Pronoun Translation(\textbf{PT}) and an overall score \textbf{TCP} that is the geometric mean of above.
Table~\ref{tab:das_tcp} shows that our model achieve comparable results as Transformer baseline and our selection strategy keeps the tokens of importance for handling discourse coherence.






\section{Analysis}
In this section, we will dive into the method and analyze some important parts and interesting phenomena. Except for extra explanation, the basic setting of all the experiments is as follows: $t=0.95$, $d_s=64$, $r=2$. Datasets are PDC.

\subsection{Sparsity Distribution}
Since the efficiency of our model totally relies on the learned topk sparse pattern, it is our major concern that to what extent the sparsity can achieve. 
As is shown in Table~\ref{tab:sparsity}, \method yields very sparse attention results. 

We also find the degree of sparsity among different modules is different. The decoder self-attention can achieve an extreme sparsity of 2\%, showing most past contexts are not crucial to the language model. While encoder self-attention only shows 10\% sparsity.
We suggest that the distribution of attention on the source side is relatively flat so the model needs more tokens.
Considering the encoder is non-autoregressive, the strong reduction of the decoder side, including cross-attention and self-attention, can significantly boost efficiency.
And even under such great sparsity, \method can still reach a comparable translation result.



\begin{table}[htbp]
    \centering
    \begin{tabular}{lccc}
        \Xhline{2\arrayrulewidth}
        Sparsity  & Enc  & Crs & Dec   \\ \hline
        Layer 0 & 10.0\%  & 3.0\% & 1.8\% \\ 
        
        Layer 3 & 9.8\% & 2.9\% & 2.7\% \\ 
        \Xhline{2\arrayrulewidth}
    \end{tabular}
    \caption{Sparsity that different attention modules can achieve. Enc, Crs, Dec refer to encoder self-attention, cross-attention and decoder self-attention respectively. Layer 0-2 and layer 3-5 share the same attention.}
    \label{tab:sparsity}
\end{table}



\subsection{Abalation Study}

Table~\ref{abaltion} shows the effects of the different modules we proposed. 

``- Top-k Selection'' means that we abandon the Top-k selection. Instead, we limit the attention range within a fixed window whose center is the query and length is 20. Though getting a lower attention cost, its quality deterioration shows that naive human prior is not robust and leads to quality drop.

``- Attention Supervision'' means that we set $\alpha$ in formula~\ref{loss} to 0, thus not constraining the consistency between the attention of the selection layer and the original layer. Consequently, the BLEU score has a large drop, showing the importance of attention supervision. And the lack of supervision might cause the failure of previous sparsity-based efficient transformers.

``- Re-parameter trick'' means that we do not use formula~\ref{trick} so that the parameters of the selection layer are only trained by attention supervision loss and do not contribute to NMT loss. The BLEU score has a small drop, showing that the re-parameter trick helps.

It achieves a comparable result but significantly raises the computation cost.

\begin{table}[htbp]
    \begin{tabular}{lcc}
        \Xhline{2\arrayrulewidth}
        \multicolumn{1}{l}{}     & Attn Cost & BLEU \\ \hline
        \multicolumn{1}{l}{\method} & 7\% & 28.04  \\
        ~- Top-k Selection   & 4\% & 26.52  \\
        ~- Attention Supervision   & 7\% & 12.94  \\
        ~- Re-parameter Trick      & 7\% & 27.58  \\
        \Xhline{2\arrayrulewidth}
    \end{tabular}
    \caption{Effects of different modules. Dynamic Selection, Attention Supervision, and Re-parameter Trick mainly contribute to the quality maintaince. Layer Sharing mainly contributes to the efficiency boost.}
    \label{abaltion}
\end{table}

\subsection{$t$ and $k$: Sparsity Effects}

The sparsity degree $k$ is also an important index. We conduct some experiments to check the relationship between the attention sum $t$ and the sparsity $k$. As is shown in Table~\ref{tab:k_and_t}, lower attention sum requirements bring more sparsity but also slightly lower BLEU. We choose $t=0.95$ as a tradeoff.



\begin{table}[h]
    \centering
    \begin{tabular}{lccc}
        \Xhline{2\arrayrulewidth}
          Threshold    & $k$    &   Attn Cost  & BLEU         \\ \hline
          $t=0.90$   &   3\%    &   5\%     & 27.90            \\
          $t=0.95$   &   5\%    &   7\%     & 28.04            \\
          $t=0.99$   &   14\%    &   16\%     & 28.22           \\
          \Xhline{2\arrayrulewidth}
    \end{tabular}
    \caption{Effects of the attention threshold and the sparsity they achieve. }
    \label{tab:k_and_t}
\end{table}

\subsection{$N$: The Longer, The More Efficient}

Since our method aims at long-range sequences, it is necessary to look into the effect of the sequence length. 
We calculate the total cost of attention (including QKV linear projection) with the sequence length.

\begin{figure}[htbp]
    \centering
    \includegraphics[width=0.5\textwidth]{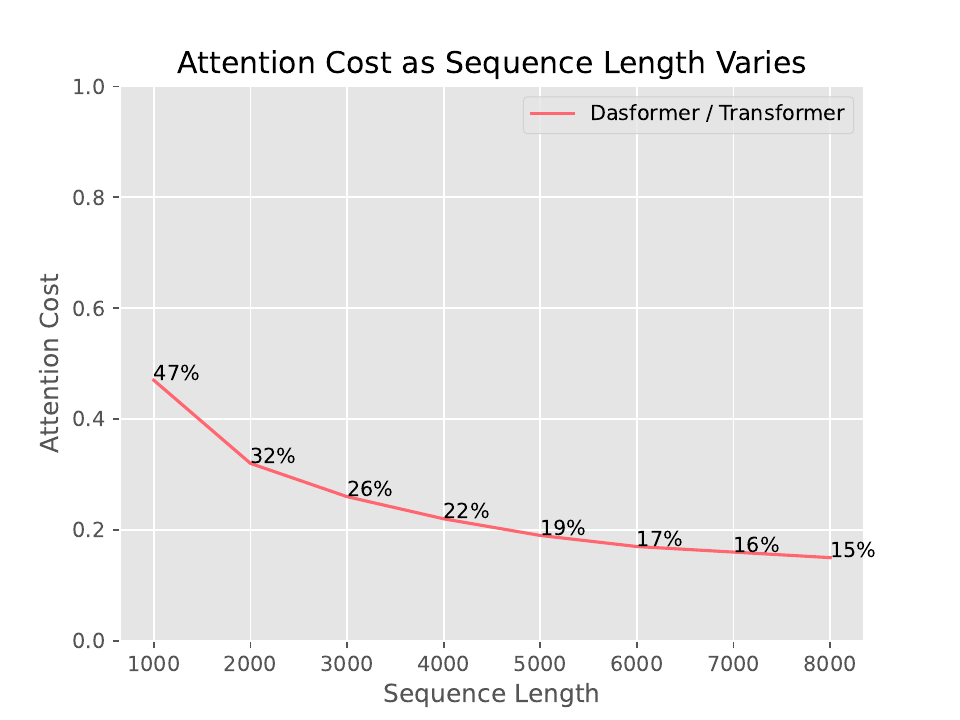}
    \caption{Attention cost (including QKV linear projection) of \method compared to original Transformer with different sequence length.}
    \label{fig:length}
\end{figure}

As is shown in Figure~\ref{fig:length}, as the sequence gets longer,
the attention cost ratio gradually decreases, which means we obtain higher and higher efficiency.
For the extremely long sequence like 8K tokens, we can lower the attention cost to 15\% (7\% if not including QKV linear projection). This shows the extraordinary potential of our methods. As the translation range gets wider and wider (e.g. a whole book or movie), \method can obtain high efficiency.

\begin{figure*}[htbp]\footnotesize
    \centering
    \subfigure[Encoder Self-attention]{
        \begin{minipage}{.32\textwidth}
            \centering
            \includegraphics[scale = 0.7]{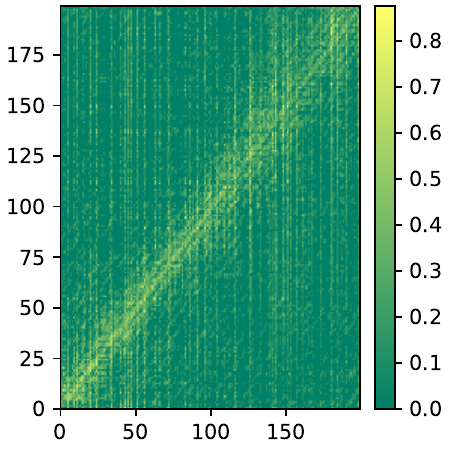}
        \end{minipage}
    }
    \subfigure[Cross-attention]{
        \begin{minipage}{.32\textwidth}
            \centering
            \includegraphics[scale = 0.7]{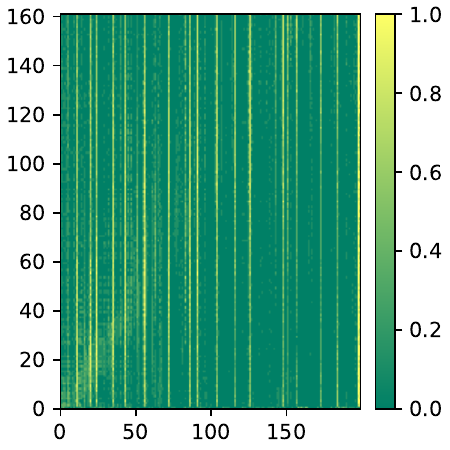}
        \end{minipage}
    }
    \subfigure[Decoder Self-attention]{
        \begin{minipage}{.32\textwidth}
            \centering
            \includegraphics[scale = 0.7]{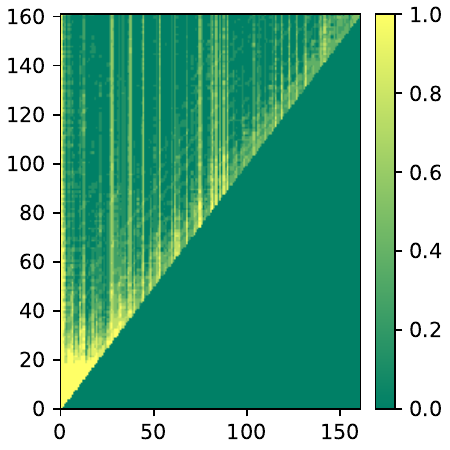}
        \end{minipage}
    }
    \caption{Visualization of all three kinds of attention. On the one hand, only a handful of tokens are necessary while most of the others are noise. On the other hand, the distribution shows some regular pattern but many attended tokens is still ruleless.}
    \label{fig:attention}
\end{figure*}

\subsection{$d_s$: U-shaped Curve with Cost}
Obviously, the low dimension sacrifices the model precision to decrease the computation cost. Therefore, it is a tradeoff to balance efficiency and performance. We conduct a series of experiments. Table~\ref{dimension} shows the efficiency and performance under different dimensions.


\begin{table}[htbp]
    \centering
    \begin{tabular}{lccc}
        \Xhline{2\arrayrulewidth}
              &  $k$                  & Attn Cost            & BLEU            \\ \hline
        Transformer  &  100\%         &   100\%             & 27.80                 \\ 
        $d_s=16$  &     24\%          &   23\%             & 24.62             \\
        $d_s=32$  &     15\%          &   16\%             & 27.88             \\
        $d_s=64$  &     5.0\%           &   7\%             & 28.04             \\
        $d_s=128$ &     4.8\%           &   9\%             & 28.05             \\
        $d_s=256$ &     4.3\%           &   13\%             & 28.06             \\
        $d_s=512$ &     5.2\%           &   21\%             & 28.20             \\ 
        \Xhline{2\arrayrulewidth}
    \end{tabular}
    \caption{Effects of different selection dimensions and the sparsity they achieve. }
    \label{dimension}
\end{table}


We find that a low dimension of 32 is enough for a coarse selection, while a dimension of 16 hurts the performance. 
Also, a lower dimension of the selection layer can bring higher sparsity $k$ which conversely raises the computation cost. 
Even if we only focus on efficiency, the lowest dimension does not mean the lowest cost. 
The attention cost goes down and then up as $d_s$ decreases. Therefore, we pick $d_s=64$ as our final setting.




\subsection{$r$: Sharing Layers Helps}
Another point is that the sparse pattern is obtained in one selection layer and applied to all of the layers within a layer group. 
We suggest that some adjacent layers share the same function so their attention can be shared together.
For example, the lower layers are expected to learn the syntactic information while the higher ones are expected to learn semantic information. So some attention distribution can be shared across the layers. 
As is shown in Table~\ref{tab:layer_sharing}, sharing layers slightly enhance $k$ and drop BLEU. We suggest that sharing too many layers limits the model capacity while not sharing results in some redundancy. Taking $r=3$ yields the best results.

\begin{table}[htbp]
    \centering
    \begin{tabular}{lcccc}
        \Xhline{2\arrayrulewidth}
                    & \multicolumn{1}{l}{$k$} & \multicolumn{1}{l}{Attn Cost} & \multicolumn{1}{l}{BLEU} \\ \hline
        \underline{012345}    & 9\% & 10\%   & 27.85       \\
        \underline{012} \underline{345}     & 5\% & 7\% & 28.04   \\
        \underline{01} \underline{23} \underline{45}  & 5\% & 8\%  & 28.12 \\
        \underline{0} \underline{1} \underline{2} \underline{3} \underline{4} \underline{5} & 5\% & 11\% &  28.26\\ 
        \Xhline{2\arrayrulewidth}
    \end{tabular}
    \caption{Attention sharing in the selection layer. The numbers sharing a underline are in the same group and share the same attention pattern.}
    \label{tab:layer_sharing}
\end{table}

\subsection{Visualization: Attention Patterns}
Figure~\ref{fig:attention} shows the sparsity patterns on encoder self-attention, cross-attention, and decoder self-attention. 

On the one hand, there exist some common characteristics, such as: 1) Most tokens prefer to attend to nearby tokens. 2) Some tokens serve as the global token that almost all tokens attend to it, which might be some punctuation. These characteristics shares the same idea with human prior\cite{DBLP:journals/corr/abs-1904-10509/Sparse,DBLP:journals/corr/abs-2004-05150/Longformer}.

On the other hand, there are also plenty of ruleless distributions, including very far tokens. We suggest that long-range context can contribute to the current token like tense or pronoun~\cite{DBLP:conf/acl/SunWZZHCL22}. These drifting attentions can not be handled by human prior while \method can well cope with it.










\section{Conclusion}
In this paper, we focus on the long-range document-level translation efficiency due to its quadratic cost growth with the length. However, previous studies suffer severe performance drops when inferring real long sequences. To address this issue,
We propose to select important tokens with lightweight attention, which is supervised by the original attention. The proposed \method effectively reduces the attention expense while successfully maintains the translation quality. It turns out that only around 5\% of attention is necessary and the attention cost can be reduced to 7\%. In the end, we achieve an overall acceleration of 20\%. 

\section*{Limitation}
The main limitation of this work is that the reduction of cost does not reflect the actual acceleration, which is influenced by linear modules
and GPU optimization. 

Linear modules include embedding layers, projection of query, key, value, and feed-forward network. 
Actually, they are the dominant bottleneck when the sequence length is short. 
We test the time cost for different modules of various input length and find that the attention modules becomes the bottleneck (over 50\%) only when the input length is over 1500 tokens. Therefore, the acceleration is relatively minor when the input is short.


GPU optimization is another important concern. First, due to the parallel computing property, a linear layer of 512 x 32 is not 8x faster than a linear layer of 512 x 512. It depends on the GPU architecture and even batch size. The more GPU cores and a small batch size result in a lower GPU utilization and a small speedup. Second, the sparse model is not as fast as a dense model in terms of GPU memory access and pre-fetching, so more memory reading cost is inevitable, which hurts the final end-to-end speedup.

\section*{Acknowledgement}
We would like to thank the anonymous reviewers for their insightful comments. Part of this work is supported by National Science Foundation of China (No. 62376116, 62176120), the Liaoning Provincial Research Foundation for Basic Research (No. 2022-KF-26-02).

\bibliographystyle{acl_natbib}
\bibliography{anthology,custom}




\end{document}